\documentclass[3p, review]{elsarticle}

\usepackage{lineno}
\usepackage{tabu}
\usepackage{algorithm}
\usepackage{algorithmic}
\usepackage{graphicx}
\usepackage{amsmath}

\usepackage{siunitx}
\usepackage{gensymb}

\usepackage[colorlinks]{hyperref}

\hypersetup{
  colorlinks=true,
  citecolor=blue,
  linkcolor=blue,
  urlcolor=blue}
\modulolinenumbers[5]

\makeatletter
\def\ps@pprintTitle{%
   \let\@oddhead\@empty
   \let\@evenhead\@empty
   \let\@oddfoot\@empty
   \let\@evenfoot\@oddfoot
}
\makeatother

\begin{document}

\begin{frontmatter}

\title{Beyond Visual Image: Automated Diagnosis of Pigmented Skin Lesions Combining Clinical Image Features with Patient Data}

\author[a,b]{Jos{\'e} G. M. Esgario}
\ead{guilherme.esgario@gmail.com}

\author[a,b,c]{Renato A. Krohling}
\ead{krohling.renato@gmail.com}

\address[a]{Federal University of Esp\'irito Santo, Vit\'oria, ES, Brazil}
\address[b]{PPGI - Graduate Program in Computer Science}
\address[c]{LABCIN, Production Engineering Department}



\begin{abstract}
Skin cancer is considered one of the most common type of cancer in several countries. Due to the difficulty and subjectivity in the clinical diagnosis of skin lesions, Computer-Aided Diagnosis systems are being developed for assist experts to perform more reliable diagnosis. The clinical analysis and diagnosis of skin lesions relies not only on the visual information but also on the context information provided by the patient. This work addresses the problem of pigmented skin lesions detection from smartphones captured images. In addition to the features extracted from images, patient context information was collected to provide a more accurate diagnosis. The experiments showed that the combination of visual features with context information improved final results. Experimental results are very promising and comparable to experts.
\end{abstract}

\begin{keyword}
\texttt{elsarticle.cls}\sep \LaTeX\sep Elsevier \sep 
\MSC[2010] 00-01\sep  99-00
\end{keyword}

\linenumbers

\end{frontmatter}

\section{Introduction}


Skin cancer is considered one of the most common type of cancer worldwide. Among the most common types of skin cancer are basal cell carcinoma, squamous cell carcinoma and melanoma. According to the \cite{who2018}, currently, between 2 and 3 million non-melanoma skin cancers and 132.000 melanoma skin cancer occur every year in the world. Melanoma is by far the most dangerous form of skin cancer, causing more than $75\%$ of all skin cancer deaths \citep{allen2016}. Early diagnosis of the disease plays an important role in reducing the mortality rate with a chance of cure greater than $90\%$ \citep{SBD}.

The diagnosis of pigmented skin lesions (PSLs) can be made by invasive and non-invasive methods. One of the most common non-invasive methods was presented by \cite{SOYER1987803}. The method allows the visualization of morphological structures not visible to the naked eye with the use of an instrument called dermatoscope. When compared to the clinical diagnosis, the use of dermatoscope by experts makes the diagnosis of PSLs easier, increasing by $10$-$27\%$ the diagnostic sensitivity \citep{mayer1997}. An investigation by \cite{vestergaard2008} evaluated the results of nine studies for the diagnosis of PSLs with naked eye and dermatoscope. The results obtained by specialists for diagnosis with the naked eye presented a sensitivity of 0.71 $(95\%$ CI\footnote{Confidence Interval} $0.59$-$0.82)$ and specificity of 0.81 $(95\%$ CI $0.48$-$0.95)$, whereas results with dermatoscope presented sensitivity of 0.90 $(95\%$ CI $0.80$-$0.95)$ and specificity of 0.90 $(95\%$ CI $0.57$-$0.98)$. To diagnose PSLs, a number of methods can be used, for example, ABCD rule \citep{NACHBAR1994}, Menzies method \citep{menzies1996} and 7-Point checklist \citep{argenziano1998}. However, diagnosis methods by dermoscopic images are qualitative or semi-quantitative in such a way that they are highly subjective and rely on the expertise of the specialist to obtain good results \citep{Binder1995}.

There is a great research interest worldwide in the development of Computer-Aided Diagnosis (CAD) systems \citep{Silveira2009} that can help not only specialists to perform more reliable diagnoses but also non-specialists to detect early malignant lesions in themselves. Most of the studies in the literature use dermoscopic images but many times someone may want a qualified opinion about a certain lesion and the only equipment available are conventional smartphone cameras. In these cases a CAD system that deals with macroscopic images would be the most suitable way.

The basic structure of a CAD system consists of the following steps: image aquisition, segmentation, feature extraction and classification \citep{masood2013}. The first step, image acquisition, can be performed by different devices such as dermatoscopes, spectroscopes, conventional digital cameras and smartphone cameras. The second step involves the artifacts removal and lesion border detection. The final steps of a CAD system are to extract a set of discriminative features and classify the PSLs images from these features.


Segmentation is an essential step in image analysis and pattern recognition in such a way that the quality of the results of a CAD system is strongly related to the quality of segmentation \citep{Pal1993}. Nevertheless, segmenting images of skin lesions can be a very difficult task because of the variety of shapes, sizes, textures and colors. In addition, there are variations such as specular reflection, brightness difference, presence of artifacts, low contrast, etc. \citep{Silveira2009}, which makes the task more complicated. Segmentation methods can be roughly grouped into the following categories: histogram thresholding, clustering, edge-based, region-based, morphological, model-based, active contours and soft computing (for example: neural network and fuzzy logic) \citep{Celebi2009}. In addition, segmentation methods can be separated into two larger groups: automatic and interactive (semi-automatic).

Among the works of skin lesions segmentation, automatic methods are the most common in the literature. Two systematic reviews were performed by \cite{Celebi2009,celebi2015} addressing the main existing methods on segmentation. Often new segmentation approaches are presented in the literature by improving on old methods or by proposing new methods that obtain better and more consistent results.

\cite{Khalid2016} presented a new segmentation technique based on Wavelet transform applied to the blue channel. It was observed in the experiments that the Wavelet transform is very useful in the removal of some artifacts. A region-based algorithm was proposed by \cite{Pennisi2016}. The algorithm performs two processes in parallel, skin region detection by threshold and segmentation. The lesion segmentation is performed by means of edge detection with the Canny method and afterwards the Delaunay triangulation is applied. The results of the two procedures are merged to generate the final lesion mask. \cite{Eltayef2017} used a hybrid method combining the Particle Swarm Optimization algorithm and the Markov Random Field method. The pre-processing step was performed by a bank of directional filters and image reconstruction methods. The approach presented great potential in automatically identifying the lesions edges.

Recently, a great research interest in  deep learning approaches and skin lesions datasets has increased, and works have been developed for skin lesion segmentation. \cite{Yuan2017} proposed a variation of a fully convolutional networks ensemble where they explicitly included information from multiple color spaces. The work was the winner of the challenge at the 2017 International Symposium on Biomedical Imaging (ISBI) \citep{ISBI2017} in the task of skin lesion segmentation. \cite{ALMASNI2018} proposed a new segmentation method via full resolution convolutional network. The proposed method learns the features of each input image pixel and does not require pre- and post-processing steps.

Besides the automatic methods, there are the interactive methods that receive this name because they require some human interaction. Among the existing interactive methods for skin lesions segmentation, the most common approaches are based on active contours and region growing that need initial seeds to perform segmentation. The user interaction level depends on the method being applied. \cite{Silveira2009}, compared six segmentation methods for dermoscopic images. The evaluated interactive methods achieved the best results.

\cite{BENSAID1996} developed a method called Semi-Supervised C-Means with the purpose of overcoming difficulties of clustering algorithms in cases where it is possible that some data of each class can be labeled. The proposed method was applied in magnetic resonance imaging and was superior compared to other methods in the literature. \cite{Surlakar2016} presented a comparative analysis of K-Means and K-Nearest Neighbors (K-NN) methods in the segmentation of histopathological images of sweat gland tissues. The results were evaluated using the mutual information metric. The authors concluded that K-NN is the best alternative for interactive segmentation of images. Recently, \cite{Luis2018} proposed an interactive segmentation algorithm for medical images, called Seeded Fuzzy C-Means (S-FCM). The method treats the seeds provided by the user as centroids and classifies the unlabeled pixels based on the similarity to the seeds. The proposed method was evaluated in several datasets, including dermoscopic image dataset.

In addition to the works focused on segmentation methods, several models of CAD systems have been presented in the literature combining different approaches of segmentation, feature extraction and classification \citep{barata2014,ferris2015,abuzaghleh2015,suganya2016,jaworek20162,bakheet2017}. The feature extraction approaches used by these systems can be divided into four classes: hand-crafted features, dictionary-based features, deep learning features and clinically inspired features \citep{fidalgo2018}. The most common ones are based on hand-crafted features inspired by the ABCD rule (asymmetry, border, color and diameter) and in texture descriptors \citep{korotkov2012}. According to the review by \cite{masood2013}, the most common classification methods used in the diagnosis of PSLs are: Artificial Neural Network, Statistical Analysis, Support Vector Machine (SVM), Decision Trees and K-NN. However, most of these approaches were applied only in dermoscopic images.

Recently, several approaches have been proposed to perform PSLs diagnoses from macroscopic images. The segmentation of macroscopic images becomes more challenging due to a variety of external factors, for instance, the difference in image brightness that is a common point on which the segmentation approaches focused on macroscopic images attempt to address \citep{wong2011,cavalcanti2013,cavalcanti20132}.

\cite{alcon2009} introduced a new system for the diagnosis of PSLs. The proposed system contains a decision-making component that combines the results of image classification and context knowledge (skin type, age, gender and part of the body) using a Bayesian network. The addition of context knowledge showed improvement in the final results. \cite{ramezani2014} developed a system for melanoma diagnosis. A set of $187$ features based on ABCD rule was extracted and to the classification step a SVM was used. \cite{chang2013} collected a total of $769$ conventional photographs and compared the classification results of the proposed CAD system with the results obtained by specialists. The results were superior to those obtained by specialists, suggesting that even with conventional images, the system has a high discriminative capacity for malignant and benign lesions. \cite{oliveira2016} developed a system capable of classify not only the PSLs but also the features of asymmetry, border, color and texture, so that the system provides the expert with both the final diagnosis and the individual features of each lesion.

Due to the increase in the smartphones processing power, some work have been carried out in order to embed complete PSLs diagnosis systems in these devices. Since smartphones have become increasingly accessible, this kind of system can contribute much in the early diagnosis of malignant lesions \citep{do2018,rat2018}.


This paper proposes a new approach for the diagnosis of PSLs from macroscopic images captured by smartphones. The proposed system combines features extracted from the image with context information about the patient in order to obtain a more accurate diagnosis. In addition, a new method of interactive segmentation is presented that takes into account the similarity of color and proximity of pixels.


The remainder of this paper is organized as follows: Section $2$ introduces the segmentation framework; Section $3$ describes the feature extraction stage; Section $4$ explains the classification method used; Section $5$ describes the proposed approach; In Section $6$ the experiments and results are presented; Finally, section $7$ presents a brief conclusion with directions for future work.

\section{Segmentation Framework}

\begin{figure}[htbp]
\centerline{\includegraphics[width=2.2in]{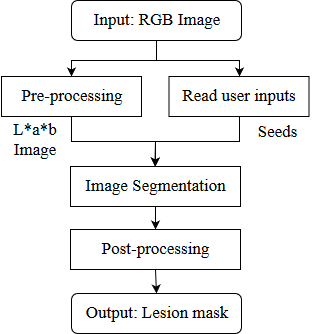}}
\caption{Block diagram of the proposed segmentation framework.}
\label{figproposedmethod}
\end{figure}

The proposed segmentation framework consists of the following steps: image acquisition, user inputs, pre-processing, image segmentation and  post-processing. In Figure \ref{figproposedmethod}, a block diagram of the proposed segmentation framework is presented. The input images were standardized with size 300x225 (4:3 ratio). The size chosen, although much smaller than the size of the original images, implies a significant reduction of the segmentation time without significant loss in the results quality. The final result returns the mask of the region of interest (ROI). The intermediate steps are detailed in the following subsections.

\subsection{User interaction}

\begin{figure}[htbp]
\centerline{\includegraphics[width=2.25in]{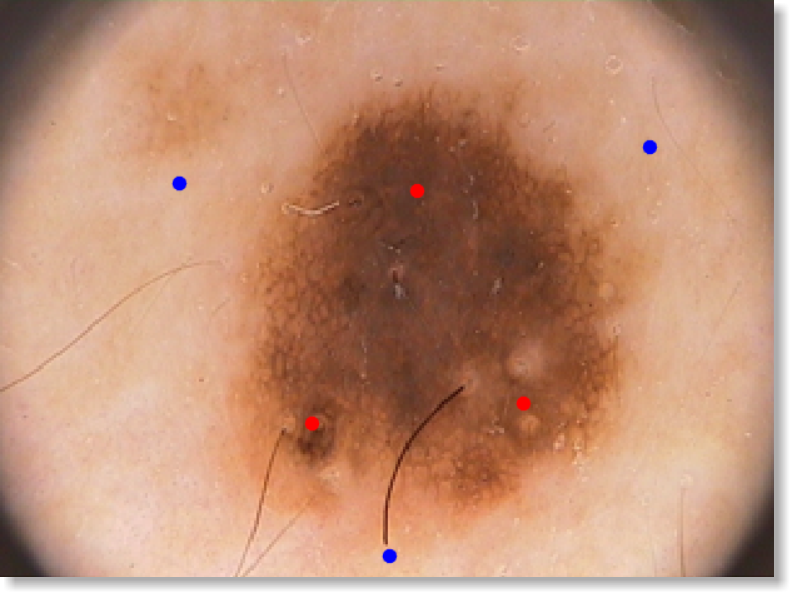}}
\caption{Example of interaction where the user selected three foreground seeds (red dots) and three background seeds (blue dots).}
\label{figseeds}
\end{figure}

User interaction is accomplished through clicks on the image. Clicks are interpreted as labeled seeds marking the lesion and the background. The only constraint of the proposed method is that at least one seed for each region must be provided by the user, i.e., the minimum user interaction to perform segmentation is two clicks. Figure \ref{figseeds} shows an example of interaction where the user selected three foreground seeds (red dots) and three background seeds (blue dots).

\subsection{Pre-processing}

\begin{figure}[htbp]
\centerline{\includegraphics[width=1.9in]{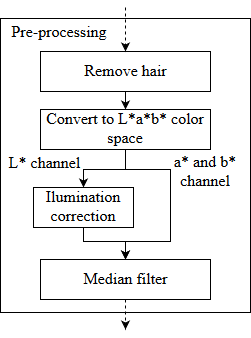}}
\caption{Block diagram of the pre-processing steps.}
\label{figpreprocessing}
\end{figure}

The motivation for applying pre-processing steps is to increase performance in the segmentation stage. Figure \ref{figpreprocessing} presents the pre-processing steps performed in this work. Figure \ref{figpreprocsteps}a was used as an example to demonstrate the results obtained in each pre-processing step. The details of each block are described in the following sections.

\subsubsection{Hair removal}

Hair is one of the most common artifacts in skin lesion imaging, its removal is essential to improve image segmentation. Most hair removal methods are based on three main steps, namely, highlights hair, segmentation and inpainting \citep{Abbas2011}. The hair was highlighted with the Laplacian of Gaussian filter with an 5x5 kernel. The hair mask was generated by fixed thresholding of $0.3$. Finally, each masked pixel of the original image was replaced by the mean value of neighboring unmasked pixels (Figure \ref{figpreprocsteps}b) in a window of $11 \times 11$ pixel size.

\subsubsection{Color space conversion}

The L*a*b* color space mimics the way human vision perceives colors making it perceptually uniform \citep{Garcia-Lamont2018}. This color space was chosen due to linear variation of hue, whose Euclidean distance, used in the segmentation method, is presented as a good metric to calculate the chromaticity difference. In addition, the image luminance (L*) is presented in a separate channel which facilitates the correction of illumination.

\subsubsection{Illumination correction}

This stage aims to normalize the images illumination, following the approach proposed by \cite{Cavalcanti2010}. The first step consists in determining a set of pixels $S$ that belongs to the skin region. In the original approach a region of $20$ x $20$ pixels of each image corner was used to estimate the illumination map. \cite{Glaister2012} used the statistical region merging method to segment skin lesion. In this work, the Otsu method \citep{otsu} is used due to its ease of implementation and speed of execution. The set of pixels $S$ is used to adjust the quadratic function given by:

\begin{equation}
    z(x,y) = P_1x^2 + P_2y^2 + P_3xy + P_4x + P_5y + P_6
\end{equation} where the adjustment of the $z(x,y)$ function is given by the choice of the coefficients $P_i (i=1,...,6)$ that minimize the error function $\epsilon$:

\begin{equation}
\epsilon = \sum_{j=1}^N{[L(S_{j,x},S_{j,y})-z(S_{j,x},S_{j,y})]^2}
\end{equation} where $S_{j,x}$ and $S_{j,y}$ are the $x$ and $y$ coordinates of the $j$-th element of the set $S$ and $N$ is the total number of pixels belonging to the skin region. In the approach proposed by \cite{Cavalcanti2010} the illumination correction is performed on channel $V$ of the $HSV$ color space. For this work the channel $L$ of the color space $L$*$a$*$b$ is used and the illumination correction is computed by the following equation:

\begin{equation}
   R(x,y) = L(x,y) - z(x,y) + mean(z)
\end{equation} where $R(x,y)$ is the final image normalized resulting from the subtraction of the original image $L(x,y)$ by the estimated illumination map $z(x,y)$. Replacing the original luminance channel  $L$ for $R$ results in an image in the color space $L$*$a$*$b$ with the normalized luminance. An example of normalized image by the illumination correction stage can be seen in Figure \ref{figpreprocsteps}c.

\subsubsection{Median filter}

\begin{figure}[htbp]
\centerline{\includegraphics[width=3.0in]{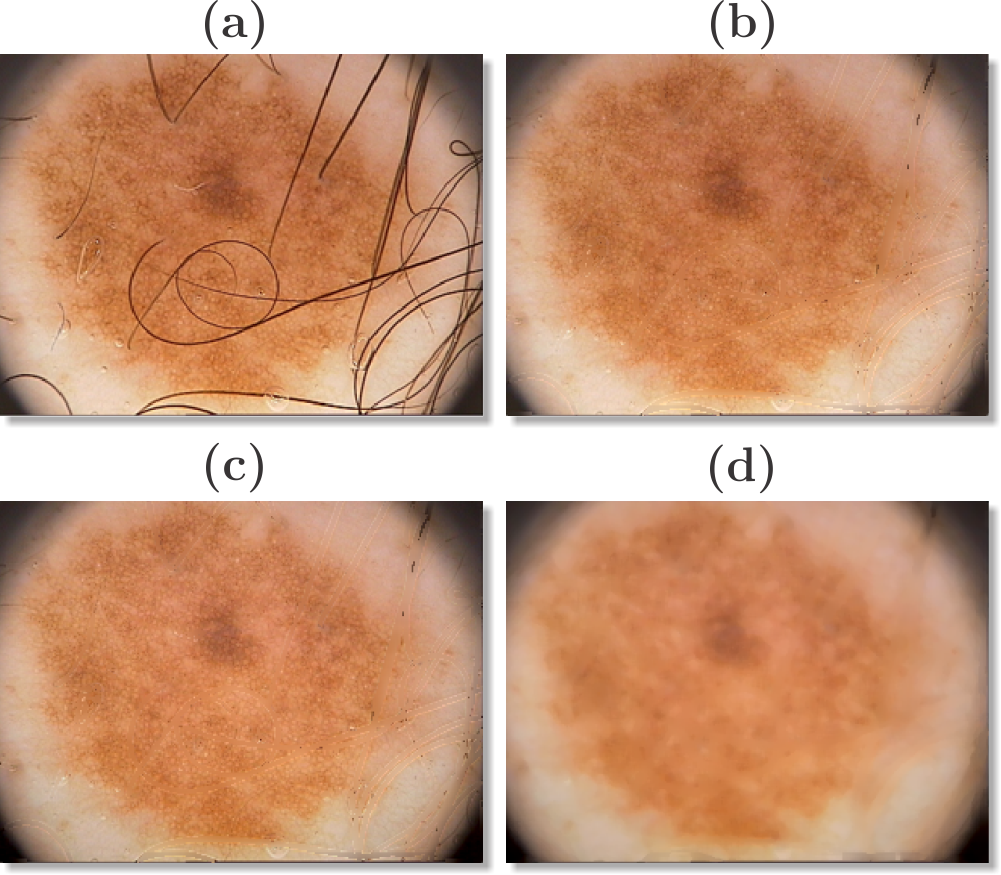}}
\caption{Pre-processing steps: (a) Original image; (b) Image after hair removal; (c) Image with normalized luminance; (d) Smoothed image by median filter. }
\label{figpreprocsteps}
\end{figure}

Lastly, a median filter with a window size of 5x5 is applied to remove small artifacts that may affect the segmentation quality. This filter was chosen for its ability to remove noise while preserving image contours \citep{oliveira2016}. Figure \ref{figpreprocsteps}d shows an example of the median filter application.

\subsection{Image Segmentation}

The image segmentation step consists of separating the pixels which represent the lesion and pixels which represent the background. Since a set of labeled seeds (pixels) are available, due to the interactive nature of the proposed method. An algorithm for pixel classification based on the nearest neighbor rule was used. This decision rule provides a simple non-parametric procedure for assigning labels to the input set based on the class label represented by the nearest neighbor \citep{Keller1985}. In order to calculate the similarity between two samples, the distance measure proposed by \cite{achanta2010} which takes into account both the similarity of color and proximity of the pixels, was used in this work. The distance measure is calculated according to

\begin{equation}
\begin{split}
d_{lab}(p,q) & = \sqrt{(p_l - q_l)^2 + (p_a - q_a)^2 + (p_b - q_b)^2} \\
d_{xy}(p,q) & = \sqrt{(p_x - q_x)^2 + (p_y - q_y)^2} \\
D(p,q) & = d_{lab}(p,q) + \frac{m}{S}d_{xy}(p,q)
\end{split}
\label{eqdist}
\end{equation} where $D(p,q)$ is equal to the sum of the $L$*$a$*$b$ distance and the $xy$ plane distance normalized between the sample pair $p$ and $q$. $S$ was defined as the diagonal size of the image $S = \sqrt{I_{rows}^2 + I_{cols}^2}$. The variable $m$ controls the importance of the spatial position in the classification of the unlabeled pixels and impacts the compactness of the final mask.

The classification is performed by assigning to the input sample $p$, the label of the $i$-th sample labeled $q_i$ that presents the shortest distance according to the following equation.

\begin{equation}
    \min{D(p,q_i)}, \ \ i=1,...,n.
    \label{eqmin}
\end{equation}

\subsection{Post-processing}

\begin{figure}[htbp]
\centerline{\includegraphics[width=3in]{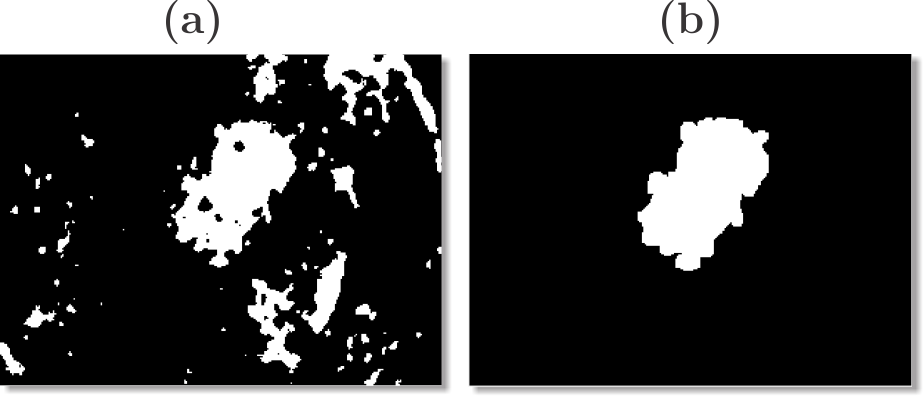}}
\caption{Post-processing example. Before (a) and after (b).}
\label{figpostprocessing}
\end{figure}

Undesired elements or holes may be present in the mask resulting from the image segmentation step. So, three post-processing steps are applied. The first step is to identify the objects in the image and check which ones have at least one seed provided by the user, otherwise the object is removed from the mask. Next, a morphological dilation operator is applied with a 3x3 sphere kernel, making the mask contour smoother. Finally, a morphological reconstruction algorithm was used to fill possible holes in the mask, i.e., fill regions of the image background that are not reachable starting from the edges \citep{soille2013}. Figure \ref{figpostprocessing} presents an example of applying the post-processing steps.

\section{Feature Extraction}

Once the ROI is identified, the next step consists in the feature extraction, i.e., relevant information is extracted from the images and information that is irrelevant to the classification step is discarded. The proposed system is based on the features analyzed by ABCD rule and texture analysis. The set of extracted features $F$ can be summarized in asymmetry ($A_{1-5}$), border irregularity ($B_{1-7}$), color ($C_{1-24}$) and texture ($T_{1-23}$) properties where $F = \{A_{1-5}, B_{1-7}, C_{1-24}, T_{1-23}\}$ has a total of $59$ features. Since the images were obtained from an unknown light source, the same lesion may present very different values in color features for different types of illumination. We used the color consistency method called Shades of Gray in order to normalize the colors of the images \citep{shades2004}.

\subsection{Asymmetry}

The purpose of this set of features is to quantify the asymmetry of the lesion, both in shape \citep{messadi2014} and color \citep{smaoui2013}.

\subsubsection{Shape asymmetry ($A_{1-3}$)}

Let $\alpha(x,y)$ be the lesion mask and $\beta(x,y)$ the new mask obtained after rotating $\alpha$ by $\ang{180}$. The asymmetry index $A_1$ is computed by the following equation

\begin{equation}
A_1 = \frac{|\alpha \cap \beta|}{|\alpha \cup \beta|}
\label{eqass}
\end{equation} where the $|.|$ operator returns the mask area.

The lesion mask is centered on the image and rotated so as to align the larger axis horizontally. A vertical cut and a horizontal cut are made in the image, resulting in 4 new masks: $\alpha_{left}$, $\alpha_{right}$, $\alpha_{top}$ and $\alpha_{down}$. Next, two new indexes $A_2$ and $A_3$ are calculated from the obtained masks in the same way as in the equation presented above.

\begin{align}
A_2 = \frac{|\alpha_{left} \cap \alpha_{right}|}{|\alpha_{left} \cup \alpha_{right}|} && A_3 = \frac{|\alpha_{top} \cap \alpha_{down}|}{|\alpha_{top} \cup \alpha_{down}|}
\end{align}

\subsubsection{Color asymmetry ($A_{4-5}$)}

To measure the asymmetry in terms of color, the sum of the Chi-square distances between the histograms of each lesion part and for each RGB component is calculated.

\begin{equation}
    D(h_1,h_2) = \sum_{c=1}^3{\sum_{i=1}^N{\frac{(h_1(i,c) - h_2(i,c))^2}{h_1(i,c) + h_2(i,c)}}}
\end{equation}

Therefore, two measures of color asymmetry are calculated: $A_4 = D(h_{left},h_{right})$ and $A_5 = D(h_{top},h_{down})$.

\subsection{Border}

To calculate border irregularity, common measures in the literature are used, among them: Compactness, Fractal Dimension, Radial Variance, Pigmentation Transition and Solidity \citep{fractal1993,bhuiyan2013,jaworek2016,lynn2017,yamunarani2018}. Besides these measures, the method of border irregularity evaluation proposed by \cite{jaworek2015} was used.

\subsubsection{Compactness ($B_1$)}

The lesion irregularity can be measured taking into account its perimeter $p$ and its area $|\alpha|$. A circle-shaped lesion has a compactness equal to $1$.

\begin{equation}
    B_1 = \frac{p^2}{4 \pi |\alpha|}
\end{equation}

\subsubsection{Fractal Dimension ($B_2$)}

The edge of the lesion mask can be modeled as a fractal curve, from this curve we can derive its fractal dimension. Fractals have their own dimension which are usually non-integer and gives the idea of how much the object fills the space in which it ``lives''. This dimension can be calculated using the box counting method. The steps that describe the calculation of the fractal dimension \citep{deviha55} are presented below.

\begin{itemize}
    \item Divide the image into regular meshes with mesh size of $r$.
    \item Counts the number of boxes that intersect at least one element of maximum level and minimum level of the binary image $N_r$.
    \item The $N_r$ value is computed for different values of $r$.
    \item The fractal dimension is represented by the slope of the curve that best fits the points $(\log{(\frac{1}{r})},\log{N_r})$.
\end{itemize}

\subsubsection{Radial Variance ($B_3$)}

The border irregularity of the lesion can be estimated by the variance of the radial distance distribution described by:

\begin{equation}
    B_3 = \frac{\frac{1}{N}\sum_{i=1}^N{(d(p_i,G)-m)^2}}{m^2}
\end{equation} where $d(p_i,G)$ is the distance from the $i$-th boundary pixel $p_i$ to the lesion centroid $G$, $m$ is the average distance $d$ and $N$ is the total boundary pixels.

\subsubsection{Pigmentation Transition ($B_{4-5}$)}

Since lesion border with abrupt variation may suggest malignancy according to the ABCD rule, this feature was used in order to describe the pigmentation transition between lesion and skin. The RGB image is transformed into a single luminance component $L(i,j)$ described by:

\begin{equation}
    L(i,j) = \frac{1}{3}(R(i,j) + G(i,j) + B(i,j))
\end{equation}

From the luminance component the magnitude of the gradient vector for each pixel belonging to the lesion edge is calculated. To describe the pigmentation transition, the mean and variance of magnitude values of the gradient $e(k)$ ($1 \leq k \leq K$ where $K$ is the total edge pixels) are calculated.

\begin{align}
B_4 = \frac{1}{K}\sum_{k=1}^K{e(k)}
&&
B_5 = \frac{1}{K}\sum_{k=1}^K{e^2(k)} - m_e^2
\end{align}

where

\begin{equation}
    e(k) = \sqrt{(\frac{\partial L_k}{\partial x})^2 + (\frac{\partial L_k}{\partial y})^2}
\end{equation}

\subsubsection{Solidity ($B_6$)}
The solidity is computed as the rate between the area of the lesion mask $\alpha$ and its convex hull area $|\alpha_{CH}|$.

\begin{equation}
    B_6 = \frac{|\alpha|}{|\alpha_{CH}|}
\end{equation}

\subsubsection{Jaworek Method ($B_7$)}

It consists of four steps \citep{jaworek2015}: skin lesion rotation, borderline function, smoothing and irregularities detection. Initially the mask is rotated so that the major axis of the lesion is parallel to the horizontal axis. Then, the bounding box of the lesion is calculated. The borderline function is generated from the distance of the lesion border pixels to the bounding box. The function is smoothed and the Jaworek border irregularity is counted as the total stationary points of the function.

\subsection{Color}

A heterogeneous color lesion may be a melanoma sign, so measures that quantify the color difference in a PSLs image are important in discriminating this sort of lesion. For this, histogram measures \citep{stoecker2009} and color variegation \citep{jaworek2016} were used.

\subsubsection{Histogram Measures ($C_{1-18}$)}

From the histograms of each channel of the RGB and HSV color spaces, the mean $\mu$, variance $\sigma$ and asymmetry $S$ are computed, where $r_i$ is the $i$-th pixel intensity level, $p(r_i)$ is the occurrence probability of intensity level $r_i$ and $L$ is the number of intensity levels. A total of $18$ histogram measures are computed.

\begin{equation}
    \mu = \sum_{i=0}^{L-1}{r_ip(r_i)}
\end{equation}

\begin{equation}
    \sigma = \sum_{i=0}^{L-1}{(r_i - \mu)^2 p(r_i)}
\end{equation}

\begin{equation}
    S = \sum_{i=0}^{L-1}{(r_i - \mu)^3 p(r_i)}
\end{equation}

\begin{center}
$C_{1-3} = \{\mu(R), \sigma(R), S(R)\}$ \\
$C_{4-6} = \{\mu(G), \sigma(G), S(G)\}$ \\ 
$C_{7-9} = \{\mu(B), \sigma(B), S(B)\}$ \\
$C_{10-12} = \{\mu(H), \sigma(H), S(H)\}$ \\
$C_{13-15} = \{\mu(S), \sigma(S), S(S)\}$ \\
$C_{16-18} = \{\mu(V), \sigma(V), S(V)\}$ \\
\end{center}

\subsubsection{Color Variegation ($C_{19-24}$)}

The color variegation is calculated as the log of the variance over mean of each image channel and color space. In this work, RGB and HSV color space are used, therefore, six new measures are computed.

\begin{equation}
    cv = \log{\frac{\sigma}{\mu}}
\end{equation}

\subsection{Texture}

Texture properties are calculated from grayscale images of the lesions, in order to quantify their structural characteristics. In this work, lesion texture properties are extracted using the Gray Level Co-occurrence Matrix (GLCM) and Gray-Level Run Length Matrix (GLRLM).

\subsubsection{GLCM ($T_{1-16}$)}

Texture properties can be computed from the GLCM, initially described by \cite{haralick1973}. The GLCM is defined as a matrix of relative frequencies $p(i,j|d,\theta)$, i.e., the frequency at which neighboring pixels in a grayscale image, separated by a distance $d$ with orientation $\theta$, occur with gray level $i$ and $j$. Contrast, correlation, energy and homogeneity measures are calculated from the GLCM and at four different angles $\theta \in \{0\degree, 45\degree, 90\degree, 135\degree\}$. As a result, $16$ new measures $T_{1-16}$ are extracted where $T_{1-4}(\theta = 0\degree)$, $T_{5-8}(\theta = 45\degree)$, $T_{9-12}(\theta = 90\degree)$ and $T_{13-16}(\theta = 135\degree)$.

\begin{equation}
    Contrast = \sum_{i=0}^{L-1}{\sum_{j=0}^{L-1}{p(i,j)(i-j)^2}}
\end{equation}

\begin{equation}
    Correlation = \sum_{i=0}^{L-1}{\sum_{j=0}^{L-1}{\frac{ijp(i,j)-\mu_x \mu_y}{\sigma_x \sigma_y}}}
\end{equation}

\begin{equation}
    Energy = \sum_{i=0}^{L-1}{\sum_{j=0}^{L-1}{p(i,j)^2}}
\end{equation}

\begin{equation}
    Homogeneity = \sum_{i=0}^{L-1}{\sum_{j=0}^{L-1}{\frac{p(i,j)}{1 + (i - j)^2}}}
\end{equation}

\subsubsection{GLRLM ($T_{17-23}$)}

With GLRLM, higher order statistical features for a given texture can be calculated. Five measures were presented in \cite{GALLOWAY1975}: short runs emphasis (SRE), long runs emphasis (LRE), run length non-uniformity (RLN), run percentage (RP) and gray level non-uniformity (GLN). Later, \cite{CHU1990} presented two new measures: low gray level run emphasis (LGRE) and high gray level run emphasis (HGRE). GLRLM matrices are of type $M_{\theta}(c,r)$ and indicate the number of gray level occurrences $c$ with runs of length $r$ (total repetitions of the primitive) and $\theta$ is the primitives inclination angle. Just as \cite{jaworek20162}, the GLRLM was calculated for four orientations $\theta \in \{0\degree, 45\degree, 90\degree, 135\degree\}$, the four resulting matrices have been added together and seven measures have been calculated from the resulting matrix.

\begin{equation}
    H =  \sum_{i=1}^{N_g}{\sum_{j=1}^{N_r}{p(i,j)}}
\end{equation}

\begin{equation}
    T_{17} = SRE = \frac{1}{H}\sum_{i=1}^{N_g}{\sum_{j=1}^{N_r}{\frac{p(i,j)}{j^2}}}
\end{equation}

\begin{equation}
    T_{18} = LRE = \frac{1}{H}\sum_{i=1}^{N_g}{\sum_{j=1}^{N_r}{j^2p(i,j)}}
\end{equation}

\begin{equation}
    T_{19} = GLN = \frac{1}{H}\sum_{i=1}^{N_g}{(\sum_{j=1}^{N_r}{p(i,j)})^2}
\end{equation}

\begin{equation}
    T_{20} = RLN = \frac{1}{H}\sum_{j=1}^{N_r}{(\sum_{i=1}^{N_g}{p(i,j)})^2}
\end{equation}

\begin{equation}
    T_{21} = RP = \frac{H}{\sum_{i=1}^{N_g}{\sum_{j=1}^{N_r}{jp(i,j)}}}
\end{equation}

\begin{equation}
    T_{22} = LGRE = \frac{1}{H}\sum_{i=1}^{N_g}{\sum_{j=1}^{N_r}{\frac{p(i,j)}{i^2}}}
\end{equation}

\begin{equation}
    T_{23} = HGRE = \frac{1}{H}\sum_{i=1}^{N_g}{\sum_{j=1}^{N_r}{i^2p(i,j)}}
\end{equation}

\section{Classification}

After feature extraction, the next and last step is to classify the PSLs. In this work we used the Extreme Learning Machine (ELM) algorithm, proposed by \cite{HUANG2006}, with regularization factor \citep{deng2009}, called Regularized Extreme Learning Machine (RELM). The RELM is used in the training of a single hidden layer feedfoward neural network (SLFN) in order to overcome the problems of slowness in training and convergence to local minima found in the backpropagation method. As a learning algorithm, an ELM offers better generalization performance, low computational cost and easy of implementation \citep{masood2013}.

Compared to SVM, which is a fairly common approach in the classification of skin cancer, the ELM presents similar generalization ability and superior computational speed \citep{liu2012}. In addition, the ELM deals naturally with multiclass problems. Since the SVM was originally developed for binary problems then there is a need to use some method to be applied to multiclass problems \citep{huang2012}.

The RELM was used instead of ELM because it presents better generalization performance and it holds strong anti-noise ability.

The following subsections describe the data pre-processing steps and the classification method used.

\subsection{Data Pre-processing}

The quality of the classification results strongly depend on the quality of the training data. Regardless of the classifier used, if the training data are incorrect or not well processed, poor models will result \citep{Kamiran2012}. Therefore, two pre-processing steps are adopted in this work: feature scaling and oversampling.

The min-max normalization method was used to rescaling the data in the interval $[0, 1]$. The features scaling general formula is given as:

\begin{equation}
    {x'}_{ij} = \frac{x_{ij} - \min{x_j}}{\max{x_j} - \min{x_j}} , i=1,\dots,m; j=1,\dots,n
\end{equation} where $x_{ij}$ represents the $j$-th feature of the $i$-th dataset sample.

Another problem that the automatic diagnosis of skin cancer directly faces is the imbalance of the dataset, due to the difficulty in collecting these data and the rarity of certain diseases. In order to overcome these difficulties, a methodology for the generation of synthetic samples, called Synthetic Minority Over-sampling Technique (SMOTE) \citep{SMOTE2002}, was applied. This method generates synthetic samples from the linear combination of the actual samples. Therefore, SMOTE was applied in all training stages and in all minority classes, so that the number of samples of all classes were the same.

\subsection{Classifier}

This work proposes the use of RELM for the classification stage of PSLs where the pre-processed data are used in the network training.

Given a set of $N$ arbitrary training samples $(x_i,t_i)$ where $x_i \in R_n$ and $t_i \in R_m$, a SLFN with $d$ hidden nodes can be modeled by the following equation:

\begin{equation}
    \sum_{i=1}^{d}\beta_i f(w_i \cdot x_j + b_i) = t_j , \ \ j=1,\dots,N.
\end{equation}

The matrix form of the previous equation can be expressed as follows:

\begin{equation}
    H\beta=T
\end{equation}

\begin{equation}
    H = \begin{bmatrix}
    f(w_1 \cdot x_1 + b_1) & \dots & f(w_d \cdot x_1 + b_d) \\
    \vdots & \ddots & \vdots \\
    f(w_1 \cdot x_N + b_1) & \dots & f(w_d \cdot x_N + b_d) \\
    \end{bmatrix}
    \label{eqh}
\end{equation} where $f(x)$ expresses the activation function of the hidden layer, $w_i = [w_{i1},w_{i2},\dots,w_{in}]^T$ is the weights vector that connects the $i$-th neuron of the hidden layer to the neurons of the input layer, $\beta_i = [\beta_{i1},\beta_{i2},\dots,\beta_{im}]^T$ is the weights vector that connects the $i$-th neuron of the hidden layer to the neurons of the output layer, and $b_i$ is the hidden neuron bias. The matrix $H$ stores the hidden layer outputs.

The first step of RELM is to calculate the $H$ matrix according to eq. \eqref{eqh}. Once the matrix $H$ have been calculated, the weights of the matrix $\beta$ can be calculated by solving the following linear system $\beta = H^\dagger T$. Considering that the number of neurons is less than the number of samples $d << N$ then the calculation of $\beta$ is given by:

\begin{equation}
    \beta = \bigg (\frac{I}{\gamma}+H^TH \bigg )^{-1} H^TT
    \label{betaeq}
\end{equation} where $I$ is the identity matrix of order $d \times d$
and $\gamma$ is the regularization factor. The ELM is a particular case of RELM when $\gamma \to \infty$.

The RELM algorithm can be summarized in three steps.

\begin{enumerate}
    \item Randomly assign input weights to $w_i$ and bias $b_i$;
    \item Compute matrix $H$ by eq. \eqref{eqh};
    \item Compute matrix $\beta$ by means of eq. \eqref{betaeq}.
\end{enumerate}

\section{Proposed approach}

\begin{figure}[htbp]
\centerline{\includegraphics[width=2.5in]{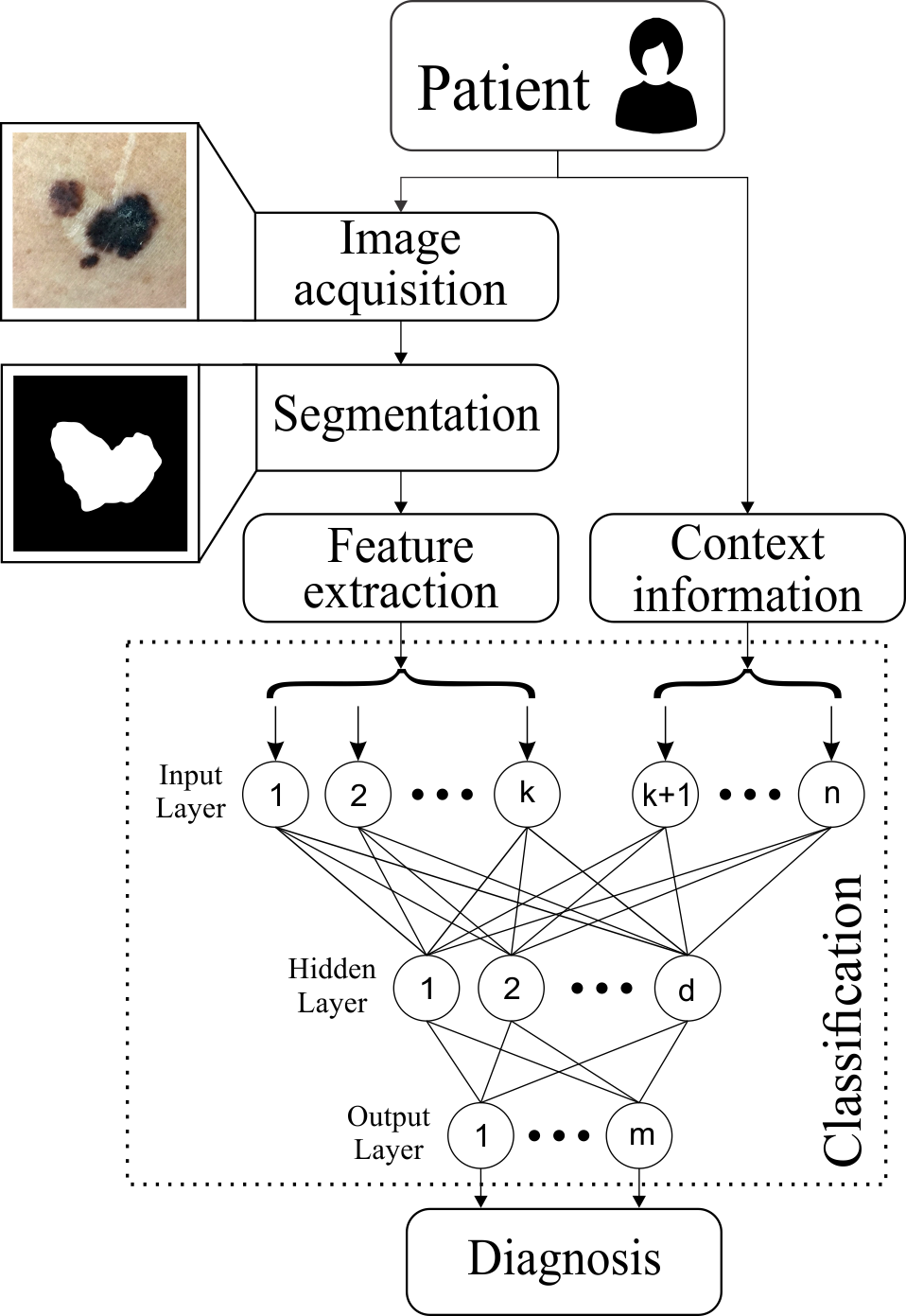}}
\caption{Block diagram of the proposed approach.}
\label{figapproach}
\end{figure}

The proposed approach encompasses all the steps mentioned above and the union of these steps produces a complete CAD system. In addition, it is proposed in this paper the combination of patient information together with the information extracted of the lesion image, the combination is performed in the classifier input vector. For datasets that do not contain context information, only image features were used. The final diagnosis returned by the system is the class referring to the highest value of the output neuron. Figure \ref{figapproach} presents a block diagram of the proposed approach.

\section{Experiments}

\subsection{Datasets}

Three datasets were used to evaluate the performance of the segmentation algorithm and the classifier. The datasets present an increasing level of difficulty and were selected in order to evaluate the performance of the algorithms under the most diverse conditions. Details of the datasets used are listed in the Table \ref{tabdatasets}.

\begin{table}[htbp]
\caption{Datasets details}
\begin{center}
\begin{tabular}{ccc}
\hline
\textbf{\textit{Name}} & \textbf{\textit{Images}} & \textbf{\textit{Lesions}} \\
\hline
PH2 & 200 & Nevus and Melanoma \\
ISBI 2017 & 2750 & Nevus, Seborrheic Keratosis and Melanoma \\
PAD-UFES & 220 & Nevus, Seborrheic Keratosis and Melanoma \\
\hline
\end{tabular}
\label{tabdatasets}
\end{center}
\end{table}

The PH2 \citep{PH2} dataset is composed of dermoscopic images acquired at Hospital Pedro Hispano under a controlled environment, with the same equipment and the same conditions. It is widely used in the literature and previous works already present excellent results for these images. Among the artifacts and challenges of this dataset we can mention the presence of hair, dark corner, specular reflection and small variation of luminosity.

The second dataset, called ISBI 2017 \citep{ISBI2017}, was developed from the collaboration of several clinical centers and is currently used in one of the the IEEE International Symposium on Biomedical Imaging (ISBI) challenges. This dataset is more challenging due to the difference in color of the images obtained by different equipments. In addition, there are the presence of various artifacts such as, colored patches, rectangular black frames, pen demarcations, presence of rulers, etc. Although this dataset has $2750$ images, only the $600$ images make up the test set proposed by the challenge are used. In addition to the images, this dataset provides the patient's age and sex information.

Finally we have a new dataset developed by the Dermatological Assistance Program (in Portuguese: Programa de Assist\^encia Dermatol\'ogica abbreviated by PAD) in cooperation with the Computing Lab. inspired by Nature (Labcin) at the Federal University of Esp\'irito Santo (UFES) that carried out a joint work to acquire images of skin lesions with smartphones. In addition to the problems encountered by dermoscopic images, the images acquired by smartphones are subject to external illumination, presence of shadows, low resolution, blurred image, etc. This dataset is made up of $88$ melanocytic nevus (NEV), $108$ seborrheic keratosis (SK) and $24$ melanoma (MEL). In addition to the images, seven context information was collected from each patient. Among them, age of each patient and six more information about the lesion that can be obtained by the following questions: ``Did it scratch?'', ``grow?'', ``hurt?'', ``change?'', ``bleed?'' and ``raise?''. The answers to the questions are summarized as ``Yes'' and ``No''.

The datasets PH2 and ISBI 2017 are common benchmarks in the literature used to evaluate the performance of new segmentation and classification methods of skin lesion images, the results for these datasets are presented in a single session. Since this is the first work performed with the PAD-UFES dataset, its results are presented in a separate session as a case study.

\subsection{Evaluation methodology}

\subsubsection{Metrics}

To evaluate the performance of the segmentation and classification methods, six common metrics in the literature were used. From the confusion matrix, five metrics were calculated: Sensitivity \eqref{mSE}, Specificity \eqref{mSP}, Accuracy \eqref{mACC}, Balanced Accuracy \eqref{mBAC} and Jaccard Index \eqref{mJA}. The sixth metric used to measure the diagnostic capacity of the proposed system is evaluated using the Receiver Operating Characteristic (ROC) curve analysis, where the area under the curve (AUC) is calculated. The ROC curve is obtained by plotting the true positive rate in function of the false positive rate by varying the decision threshold \citep{bradley1997}. The AUC is calculated as the integral of the ROC curve.

\begin{equation}
    SE = \frac{TP}{TP + FN}
    \label{mSE}
\end{equation}

\begin{equation}
    SP = \frac{TN}{TN + FP}
    \label{mSP}
\end{equation}

\begin{equation}
    AC = \frac{TP + TN}{TP + FP + TN + FN}
    \label{mACC}
\end{equation}

\begin{equation}
    BAC = \frac{SE + SP}{2}
    \label{mBAC}
\end{equation}

\begin{equation}
    JI = \frac{TP}{TP + FP + FN}
    \label{mJA}
\end{equation}

The variables TP, TN, FP and FN are abbreviations of True Positive, True Negative, False Positive and False Negative, respectively. At the pixel level these variables are defined according to the Table \ref{tabpixels} for the segmentation results where $1$ is a foreground pixel and $0$ is a background pixel.

\begin{table}[htbp]
\caption{TP, TN, FP and FN at the pixel level}
\begin{center}
\begin{tabular}{ccc}
\hline
& \textbf{\textit{Predicted Pixel}} & \textbf{\textit{Actual Pixel}} \\
\hline
TP & 1 & 1 \\
TN & 0 & 0 \\
FP & 1 & 0 \\
FN & 0 & 1 \\
\hline
\end{tabular}
\label{tabpixels}
\end{center}
\end{table}

The evaluation metrics above naturally deal with classification results of binary problems, however, since the problem of classifying PSLs is often formulated as a multiclass problem, the use of such metrics may not be intuitive. In multiclass problems, the calculation is performed as the average of each per-class metric. For more details refer to the work of \cite{hossin2015}.

\subsubsection{Segmentation methods comparison}

Unlike automatic methods, evaluating the performance of interactive methods is a more complicated task due to the need of user interaction to provide the labeled seeds to the algorithm. In this work the algorithm \ref{algperf} is proposed in order to simulate expert inputs. So, some assumptions are considered. Assuming the user is an expert then he/she will be able to:

\begin{enumerate}
    \item Insert the most suitable seed quantity;
    \item Choose the best spatial position for each seed.
\end{enumerate}

Based on these premises, the pseudo-code as shown in the algorithm \ref{algperf}, works as follows. At each iteration of the internal $for$ loop from $i=1$ to $maxEvaluation$, a set of random image seeds is selected using the ground truth mask present in the datasets. The segmentation procedure is then performed with the selected seeds and their result is stored in the vector $E$. The external $for$ loop stores the best result obtained with the random seeds for a number of $n$ points from $2$ to $maxInputSeeds$. The input $I$ refers to the original lesion image, $M_{GT}$ is the ground truth mask of the lesion, $M_{SR}$ is the segmentation result mask, $maxInputSeeds$ was set to $10$ and finally $maxEvaluation$ set to $20$. The algorithm returns the values of the evaluation metrics $F_{best}$ for the solution that obtained the best Jaccard Index.

\begin{algorithm}[H]
    \caption{Interactive method performance evaluation}
    \label{algperf}
    \hspace*{\algorithmicindent} \textbf{Input:} $I, M_{GT}$ \\
 	\hspace*{\algorithmicindent} \textbf{Output:} $F_{best}$
    \begin{algorithmic}[1]
        \FOR{$n=2$ \TO maxInputSeeds}
        	\STATE $n_{fg} \gets$ floor($n/2$)
        	\STATE $n_{bg} \gets$ ceil($n/2$)
        	\FOR{$i=1$ \TO maxEvaluation}
        	    \STATE $S \gets$ SelectRandomSeeds($M_{GT}$, $n_{fg}$, $n_{bg}$)
        	    \STATE $M_{SR} \gets$ Segmentation($I$, $S$)
        	    \STATE $E(i) \gets$ ComputeEvaluationMetrics($M_{SR}$, $M_{GT}$)
        	\ENDFOR
    	\STATE $F(n-1) \gets$ max($E$)
    	\ENDFOR
    	\STATE $F_{best} \gets$ max($F$)
    	\RETURN $F_{best}$
    \end{algorithmic}
\end{algorithm}

The proposed method, called Interactive Segmentation based on Nearest Neighbor (ISNN), is compared with recent works in the literature that obtained the best results in PH2 and ISBI 2017 datasets. The results of the automatic methods presented in this work were collected directly from the literature, and the S-FCM interactive method was implemented and evaluated using the proposed algorithm \ref{algperf}. The grouping threshold parameter of the S-FCM method was set to $l=30\%$. The pre- and post-processing steps used in the proposed algorithm were maintained in the evaluation of the S-FCM method, in such a way that only the image segmentation block was changed, providing the same structure of the segmentation framework for both methods.

\subsection{Segmentation Results}

The performance results of the algorithms for the datasets PH2 and ISBI 2017 are presented in the Table \ref{results}. All results were calculated as the average of the individual results for each image and the best results are shown in bold. 

\begin{table}[htbp]
\caption{Segmentation Results}
\begin{center}
\begin{tabular}{ccccc}
\hline
\textbf{Method} & \textbf{JI(\%)} & \textbf{SE(\%)} & \textbf{SP(\%)} & \textbf{AC(\%)} \\
\hline
\multicolumn{5}{c}{\textbf{\textit{PH2}}}\\
\hline
ISNN$_{C=0.1}$ & \textbf{93.43} & \textbf{97.16} & 97.12 & \textbf{98.05} \\
ISNN$_{C=0.25}$ & 93.36 & 97.13 & 97.09 & 97.99 \\ 
ISNN$_{C=0.5}$ & 92.67 & 96.62 & 96.80 & 97.73 \\ 
S-FCM & 92.55 & 96.80 & 95.81 & 97.60 \\
Al-masni et al. \cite{ALMASNI2018} & 84.79 & 93.72 & 95.65 & 95.08 \\
Eltayef et al. \cite{Eltayef2017} & --- & 93.88 & \textbf{97.58} & 94.74 \\
Khalid et al. \cite{Khalid2016} & --- & 93.87 & 94.57 & 94.72 \\
Pennisi et al. \cite{Pennisi2016} & --- & 80.24 & 97.22 & 89.66 \\ \hline
\multicolumn{5}{c}{\textbf{\textit{ISBI 2017}}} \\
\hline
ISNN$_{C=0.1}$ & \textbf{88.36} & \textbf{93.29} & 96.96 & \textbf{97.14} \\
ISNN$_{C=0.25}$ & 87.66 & 93.10 & 96.82 & 96.90 \\
ISNN$_{C=0.5}$ & 86.20 & 92.97 & 96.27 & 96.48 \\
S-FCM & 87.53 & 92.43 & 96.41 & 96.65 \\
Al-masni et al. \cite{ALMASNI2018} & 77.11 & 85.40 & 96.69 & 94.03 \\
Yuan and Lo \cite{Yuan2017} & 76.50 & 82.50 & 97.50 & 93.40 \\
Berseth \cite{MattBerseth2017} & 76.20 & 82.00 & 97.80 & 93.20 \\
Bi et al. \cite{Bi2017} & 76.00 & 80.20 & \textbf{98.50} & 93.40 \\
\hline
\end{tabular}
\label{results}
\end{center}
\end{table}

The experimental performance results show that the interactive methods are superior to the automatic methods, presenting a considerable difference, especially in the Jaccard Index. The proposed method ISNN with $C=0.1$ obtained the best results on all datasets and for almost all evaluation measures. The values of the variable $C$ have a direct impact on the final results, this is due to the increase or decrease in the importance of the pixels spatial position at the moment of classification, such that the higher the $C$ value, the greater the influence of the pixels on the final result, making the mask more compact. The results present an inferior performance by increasing the $C$ value. However, it was noticed in the tests that high values of $C$, like $C = 0.5$, have good application in cases that there is low color contrast between the lesion and the skin. In these cases only the color information is not enough to perform a good segmentation.

\begin{figure}[htbp]
\centerline{\includegraphics[width=3.0in]{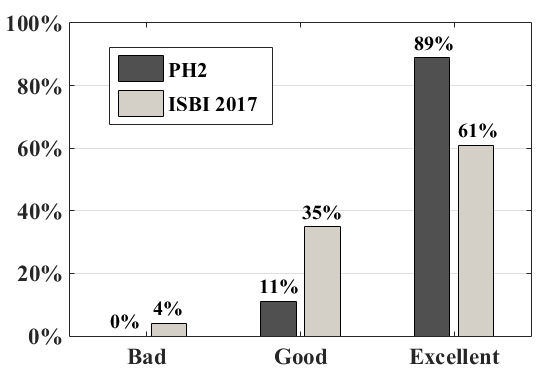}}
\caption{Percentage of images that got bad, good, and excellent results for each dataset based on the Jaccard Index.}
\label{figdiscreteja}
\end{figure}

A more detailed analysis of the ISNN$_{C=0.1}$ method results was performed based on the Jaccard Index where the final segmentation result for each image was classified as:
Bad (JA$<$0.65), Good (JA$\geq$0.65 and JA$<$0.9) or Excellent (JA$\geq$0.9). The relative frequencies of these results for each dataset are shown in a bar chart in Figure \ref{figdiscreteja}. By analyzing the bar chart, the difference in the difficulty level of the PH2 and ISBI datasets is clear, so that the percentage of excellent segmentations decreases abruptly.

The computational time was evaluated for each stage of the proposed segmentation framework. All experiments were developed and tested in the MATLAB 2015a development environment on a computer with an Intel Core i5 processor, 8GB of RAM and Microsoft Windows 10 Operating System. The computational time in seconds was calculated in the steps subsequent to the user inputs, the results are presented in the Table \ref{tabtime}.

\begin{table}[htbp]
\caption{Computational time}
\begin{center}
\begin{tabular}{ccccc}
\hline
& \textbf{Pre-processing} & \textbf{Segmentation} & \textbf{Post-processing} & \textbf{Total} \\
\hline
time(s) & 0.197 & 0.098 & 0.014 & 0.309 \\
\hline
\end{tabular}
\label{tabtime}
\end{center}
\end{table}

The framework presents a fast execution time. Pre-processing is the most time consuming step, most due the hair removal and illumination correction methods.

\subsection{Classification Results}

In order to evaluate the discriminative capacity of the extracted features in the classification of PSLs, the proposed system was tested in the same datasets. The input vector of the RELM network is formed by the union of the extracted features with the context information available in each dataset. In order to evaluate the proposed system, the leave-one-out cross validation method was used, since the dataset is unbalanced and small this method is presented as more appropriate. In addition, the result was computed as the average of $50$ runs since the RELM weights are initialized at random.

\subsubsection{PH2}

\begin{figure}[htbp]
\centerline{\includegraphics[width=3.0in]{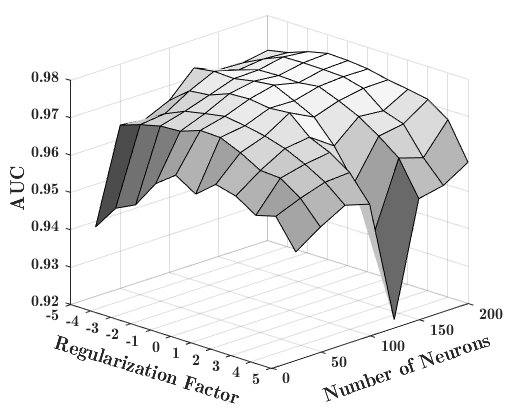}}
\caption{Sensitivity analysis of the RELM parameters.}
\label{figsurf}
\end{figure}

Initial experiments were performed for the PH2 dataset. Since there is no well defined methodology for adjusting the RELM parameters a preliminary study of the parameters sensitivity analysis was carried out. The RELM has two parameters, number of neurons $d$ and regularization factor $\gamma$. A series of tests were performed for the combinations of the parameter ranges $d = [25,50,\dots,200]$ and $\gamma = [-5,-4,\dots,5]$. The AUC value was calculated for each pair of parameters and a surface curve (Figure \ref{figsurf}.) was generated to visualize the impact of the parameters on the final result. 

A reading of the surface chart suggests that values from $d > 100$ for the interval $\gamma = [-3,0]$ presented the best results. Therefore, the parameters $d=150$ and $\gamma=-2$ were selected empirically for all subsequent experiments.

The experimental results obtained in the PH2 dataset are shown in the Table \ref{resph2} along with results from recent work in the literature where the best results are highlighted in bold. The proposed method obtained the best specificity value (SP) and a median sensitivity value (SE). Clearly the extracted features were able to perform a good classification of the PH2 dataset, making the proposed method competitive in relation to current approaches.

\begin{table}[htbp]
\caption{PH2 dataset classification results}
\begin{center}
\begin{tabular}{cccccc}
\hline
\textbf{Method} & \textbf{AUC} & \textbf{SE} & \textbf{SP} & \textbf{AC} & \textbf{BAC} \\ \hline
Barata \cite{barata2014} & - - - & 0.960 & 0.800 & 0.823 & 0.880 \\
Lei Bi \cite{bi2016} & 0.938 & 0.875 & 0.931 & 0.920 & 0.903 \\
Omar \cite{abuzaghleh2015} & - - - & \textbf{1.000} & 0.915 & 0.932 & \textbf{0.958} \\
Pennisi \cite{Pennisi2016} & - - - & 0.935 & 0.871 & \textbf{0.936} & 0.903 \\
\textbf{Proposed} & \textbf{0.974} & 0.902 & \textbf{0.933} & 0.926 & 0.917 \\
\hline
\end{tabular}
\label{resph2}
\end{center}
\end{table}

\subsubsection{ISBI 2017}

The ISBI 2017 is a multiclass dataset, however, the challenge separated the classification of lesions into two binary problems: (1) ``Melanoma against seborrheic keratosis and nevus'' and (2) ``Seborrheic keratosis against melanoma and nevus''. In addition to the features calculated from the images, patients context information like sex and age were used to classify the images.
The results obtained are presented in the Table \ref{resisic} and were compared with the ISBI 2017 winner \citep{kazuhisa2017}. It is worth mentioning that the ISBI 2017 winner used an approach based on an Ensemble of Convolutional Neural Networks as well as an external dataset. The development of an approach such as that of \cite{kazuhisa2017} demands a lot of images and high computational cost. Despite the inferior results obtained by our approach, the average accuracy result was close to the winner of the ISBI 2017 and even better in the classification of seborrheic keratosis.

\begin{table}[htbp]
\caption{ISBI 2017 dataset classification results}
\begin{center}
\begin{tabular}{cccccc}
\hline
\textbf{Method} & \textbf{AUC} & \textbf{SE} & \textbf{SP} & \textbf{AC} & \textbf{BAC} \\ \hline
\multicolumn{6}{c}{Melanoma vs Rest} \\ \hline
ISBI 2017 Winner & \textbf{0.868} & \textbf{0.735} & \textbf{0.851} & \textbf{0.828} & \textbf{0.793} \\
\textbf{Proposed} & 0.806 & 0.695 & 0.780 & 0.765 & 0.737 \\ \hline
 \multicolumn{6}{c}{Seborrheic keratosis vs Rest} \\ \hline
ISBI 2017 Winner & \textbf{0.953} & \textbf{0.978} & 0.773 & 0.803 & \textbf{0.876} \\
\textbf{Proposed} & 0.883 & 0.795 & \textbf{0.813} & \textbf{0.810} & 0.804 \\ \hline
\multicolumn{6}{c}{Average} \\ \hline
ISBI 2017 Winner & \textbf{0.911} & \textbf{0.856} & \textbf{0.812} & \textbf{0.816} & \textbf{0.834} \\
\textbf{Proposed} & 0.845 & 0.745 & 0.797 & 0.788 & 0.771 \\ \hline
\end{tabular}
\label{resisic}
\end{center}
\end{table}

\subsection{Case study}

The case study was performed with the dataset of macroscopic images, PAD-UFES. All the experiments performed in this study follow the same methods and parameters used in the results evaluation of the PH2 and ISBI 2017 benchmarks. First the results of segmentation and then the classification results are presented. \\

\subsubsection{Segmentation results}

The performance results of algorithms ISNN and S-FCM for the dataset PAD-UFES are presented in Table \ref{results}. All results were calculated as the average of the individual results for each image.

\begin{table}[htbp]
\caption{PAD-UFES - Segmentation Results}
\begin{center}
\begin{tabular}{ccccc}
\hline
\textbf{Method} & \textbf{JI(\%)} & \textbf{SE(\%)} & \textbf{SP(\%)} & \textbf{AC(\%)} \\
\hline
ISNN$_{C=0.1}$ & \textbf{84.08} & \textbf{94.02} & 98.50 & \textbf{98.21} \\
ISNN$_{C=0.25}$ & 83.02 & 93.65 & 98.38 & 98.07 \\ 
ISNN$_{C=0.5}$ & 81.00 & 93.68 & 97.93 & 97.69 \\ 
S-FCM & 82.88 & 91.35 & \textbf{98.66} & 98.06 \\ \hline
\end{tabular}
\label{results}
\end{center}
\end{table}

The proposed method ISNN with $C=0.1$ also presented the best result on this dataset. The segmentation quality were evaluated in terms of the Jaccard index and classified as: Bad (JA$<$0.65), Good (JA$\geq$0.65 and JA$<$0.9) and Excellent (JA$\geq$0.9). As a result, the percentages were: $11\%$, $51\%$ and $38\%$ respectively.
Despite the difficulties encountered in segmenting macroscopic images, only $11\%$ of the segmented images were rated as Bad.

\begin{figure}[htbp]
\centerline{\includegraphics[width=3.5in]{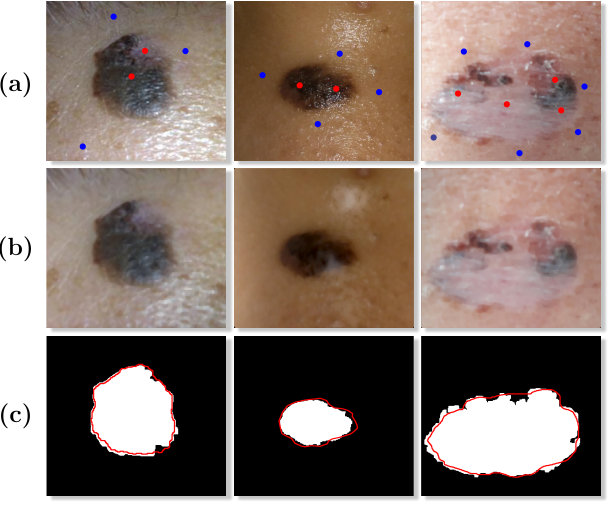}}
\caption{Application examples of the proposed method: (a) Original image with user inputs; (b) Pre-processed image; (c) Segmentation result (white) and \textit{Ground Truth} (red contour). }
\label{figresults}
\end{figure}

Segmentation examples are presented in Figure \ref{figresults} in order to exemplify the behavior of the method. Figure \ref{figresults}(a) consists of the original image with seeds provided by the user. Figure \ref{figresults}(b) shows the image after the pre-processing step and Figure \ref{figresults}(c) shows the mask resulting from the segmentation step compared to the Ground Truth mask generated by the expert. \\

\subsubsection{Classification results}

\begin{figure}[htbp]
\centerline{\includegraphics[width=2.7in]{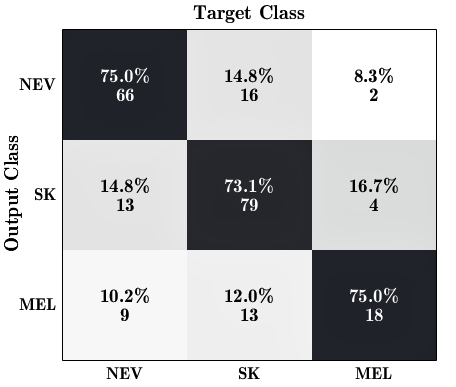}}
\caption{Confusion matrix describing the performance for the PAD-UFES dataset.}
\label{figcm}
\end{figure}

Initial experiments of multiclass classification were performed with the PAD-UFES dataset to verify the benefits of combining image features with context information. The classification results obtained with the proposed method are shown in the Table \ref{respad1}. The \textit{Experiment} column describes which features were used and the $n_f$ column represents the number of features used in each experiment. The confusion matrix for one of the experiment runs using image features and patient context information is shown in Figure \ref{figcm}.

\begin{table}[htbp]
\caption{PAD-UFES dataset classification results}
\begin{center}
\begin{tabular}{cccccc}
\hline
\textbf{Experiment} & $\textbf{\textit{n}}_f$ & \textbf{SE} & \textbf{SP} & \textbf{AC} & \textbf{BAC} \\ \hline
Image features & 59 & 0.664 & 0.816 & 0.640 & 0.740 \\
Context information & 7 & 0.684 & 0.836 & 0.692 & 0.760 \\
Combined & 59+7 & \textbf{0.744} & \textbf{0.872} & \textbf{0.753} & \textbf{0.808} \\ \hline
\end{tabular}
\label{respad1}
\end{center}
\end{table}

The results of the combination of features shown in Table \ref{respad1} presented an accuracy difference of $ 11.3 \% $ in relation to the experiment with the image features, it turns out that the expressive gains are due to the low correlation of the context information with the image features so that its addition to the features vector adds important information for the PSLs classification. The confusion matrix presented in Figure \ref{figcm} shows that the major errors made by the classifier are related to the SK class whereas the NEV and MEL classes appear easier to distinguish.

\begin{figure*}[htbp]
\centerline{\includegraphics[width=7.2in]{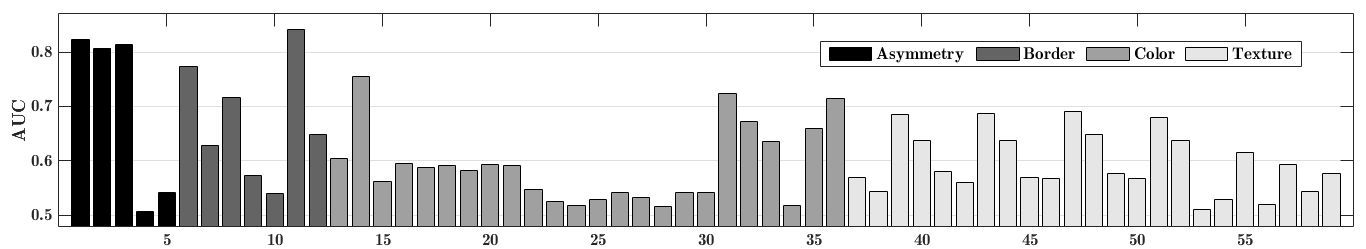}}
\caption{Univariate analysis of features based on the AUC metric for classification of melanoma and non-melanoma.}
\label{figfeatanal}
\end{figure*}

Since we are interested in detecting melanoma, the problem can be converted to a binary classification problem by grouping the seborrheic keratosis and nevus classes into a single non-melanoma class. An univariate analysis based on the AUC metric was performed to individually evaluate the $59$ features in the classification of the melanoma and non-melanoma classes. The univariate analysis result can be visualized in Figure \ref{figfeatanal}. In this case, we highlight $9$ features that obtained an AUC value higher than $0.7$, being: Shape Asymmetry ($A_{1-3}$), Compactness ($B_1$), Radial Variance ($B_3$), Solidity ($B_6$), Red Channel Variance $R$ ($C_2$) and Variegation of Channels $R$ ($C_{19}$) and $S$ ($C_{24}$). Classification results between melanoma and non-melanoma are presented in the Table \ref{respad2}. The experiments were performed for different isolated sets of features and their combination.

\begin{table}[htbp]
\caption{PAD-UFES dataset classification results between melanoma and non-melanoma.}
\begin{center}
\begin{tabular}{ccccccc}
\hline
\textbf{Experiment} & $\textbf{\textit{n}}_f$ & \textbf{AUC} & \textbf{SE} & \textbf{SP} & \textbf{AC} & \textbf{BAC} \\ \hline
Image features & 59 & 0.838 & 0.652 & 0.844 & 0.823 & 0.748 \\
Image features* & 9 & 0.862 & 0.671 & 0.850 & 0.831 & 0.761 \\
Context information & 7 & 0.853 & 0.667 & 0.803 & 0.788 & 0.735 \\
Combined & 59+7 & 0.904 & 0.726 & \textbf{0.902} & \textbf{0.883} & 0.814 \\
Combined* & 9+7 & \textbf{0.928} & \textbf{0.777} & 0.894 & 0.881 & \textbf{0.835} \\ \hline
\multicolumn{6}{l}{* Experiments with the best image features.}
\end{tabular}
\label{respad2}
\end{center}
\end{table}

The experimental results presented in Table \ref{respad2} shows that the performance of the classifier is significantly improved by the combination of image features and context information. In addition, the selection of the best image features also improved the results with an expressive reduction in the total of features used. Based on the results obtained by our system ($SE$:$0.778$ and $SP$:$0.894$) and the results of clinical diagnosis to the naked eye presented by \cite{vestergaard2008} ($SE$:$0.71, \ 95\%$ CI $0.59$-$0.82$ and $SP$:$0.81, \ 95\%$ CI $0.48$-$0.95$) we can observe that the diagnostic capacity of the proposed system is comparable to an expert.

As pointed out by \cite{haenssle2018} and reinforced by \cite{brinker2018} deep learning (e.g., convolutional neural networks) alone does not solve the problem of skin cancer diagnostic yet and an improvement in classification performance could be achieved by adding clinical data as inputs to the classifiers. As far as we know, our approach combining image features with patient context information is one of the first in this direction showing that patient information is essential in the classification process similar to the decision making of dermatologists. The dataset collected in PAD-UFES with smartphone images and clinical data and the source code are available from authors upon request. 

\section{Conclusion}

This work presented a new system for classification of pigmented skin lesions capable of dealing with dermoscopic and macroscopic images. In this study a new method of interactive segmentation was presented in order to overcome the difficulties of automatic methods. The Regularized Extreme Learning Machine algorithm was used to train a single hidden layer feedfoward neural network for the classification of pigmented skin lesions using a set of $59$ features based on ABCD rule and texture analysis. Since macroscopic images may not be very informative due to low resolution, blur and the presence of many artifacts, we have investigated the use of patient context information in addition to the extracted features from the images. The classification  results of pigmented skin lesions indicated that the performance of the proposed system is comparable to that of an expert by the naked eye.

Future work shall investigate the system's ability to classify a larger number of classes so that the CAD system can assist specialists in diagnosing a wider range of lesions. In addition, new features that might increase system performance should be investigated.

\section*{Acknowledgements}

XXX thanks the XXX for funding this study - Finance Code XXX. XXX would like to thank the XXX and the local Agency of the state of XXX for financial support under grant No. XXX and No. XXX, respectively.

\bibliographystyle{model5-names.bst}\biboptions{authoryear}
\bibliography{references}

\end{document}